# Analysis of different disparity estimation techniques on aerial stereo image datasets

Ishan Narayan[1,2*] and Shashi Poddar[1,2]

[1*]Academy of Scientific and innovative Research, AcSIR Headquarters
CSIR-HRDC Campus, Kamla Nehru Nagar, Ghaziabad, 201002, Uttar Pradesh, India.
[2]IMCS, Central Scientific Instrument Organization, CSIO, Chandigarh, 160030, Chandigarh, India.

Contributing authors: shashipoddar@csio.res.in;

# 1 ABSTRACT

With the advent of aerial image datasets, dense stereo matching has gained tremendous progress. This work analyses dense stereo correspondence analysis on aerial images using different techniques. Traditional methods, optimization based methods and learning based methods have been implemented and compared here for aerial images. For traditional methods, we implemented the architecture of Stereo SGBM while using different cost functions to get an understanding of their performance on aerial datasets. Analysis of most of the methods in standard datasets has shown good performance, however in case of aerial dataset, not much benchmarking is available. Visual qualitative and quantitative analysis has been carried out for two stereo aerial datasets in order to compare different cost functions and techniques for the purpose of depth estimation from stereo images. Using existing pre-trained models, recent learning based architectures have also been tested on stereo pairs along with different cost functions in SGBM. The outputs and given ground truth are compared using MSE, SSIM and other error metrics.
keywords: Stereo images, Depth estimation, Semi-global block matching, Unmanned aerial vehicle, Learning based methods

# 2 Introduction

Unmanned aerial vehicles (UAVs) are widely used for tasks such as mapping, surveillance, farming, and remote sensing. However, a key challenge remains: enabling autonomous landing in unknown areas. While advancements in simultaneous localization and mapping (SLAM) have improved UAV navigation in GPS-denied environments, autonomous landing still requires knowledge of the landing surface. In emergency or isolated areas, UAVs may need to land on varied surfaces, such as rocky, flat, or slanted terrain. Several algorithms exist for autonomous landing site selection, but determining surface properties like flatness and inclination often relies on stereo vision to generate dense depth maps. Unfortunately, existing disparity estimation techniques sometimes yield inaccurate depth estimates in UAV images, making it challenging to assess surface characteristics. Stereo vision, which uses two cameras to create 3D models by analyzing disparities between images, has been extensively studied, but existing benchmarks still need to include aerial images. These images present unique challenges, such as low resolution, low-texture areas, and varied depth distribution. The low texture is often due to top-down views of buildings or vegetation, making feature matching difficult.

This study evaluates several dense disparity estimation schemes on aerial stereo datasets for real-time applications like landing site selection. Initial experiments revealed that SGBM-based algorithms are commonly used, while optimization-based methods, though accurate, are too slow for real-time use. Deep learning architectures were also compared for their performance on aerial imagery. To our knowledge, such a comparison has not been done for UAV-specific images. With the growing use of autonomous UAVs, this study aims to guide future depth estimation models for UAV imagery.

The paper is structured as: Section 2 covers different disparity estimation schemes, Section 3 details SGBM and deep learning algorithms, Section 4 presents experiments on the WHU Stereo and Mid Air datasets, and Section 5 concludes the paper.

# 3 Literature Review

Dense depth map estimation is a computer vision approach that aims at obtaining the depth of an image point given images from two rectified cameras with known baseline distance. Although there is a large number of disparity estimation algorithms existing in the literature, it is still a challenging task to handle occlusions, texture-less regions and discontinuities in the images [2]. Most of these algorithms have been tested on synthetic and real-world images captured from the front and very less experiments have been done on images captured from the top view as is the case with aerial images [3]. The images captured from aerial platform face some challenges like large disparity search space, bigger occlusions and varied distribution. It is thus necessary to study the aerial stereo images holistically and benchmark the performance of different classes of disparity estimation algorithms on them. Dense depth map estimation algorithms can be divided into global approaches, local, and semi-global depending on cost aggregation methodology used [2]. With rise in usage of deep learning approaches for depth map estimation, these can also be classified as traditional or learning based methods [4].

## 3.1 Traditional Depth estimation techniques

Traditionally, the four key processes involved in depth estimation, the steps include cost calculation, cost aggregation, disparity computation, and disparity refinement. Various cost functions, such as absolute difference (AD), squared difference (SD), normalized cross-correlation (NCC), mutual information (MI), and non-parametric transforms like rank and census transforms, are used to match left and right pixels. [6]. Different cost functions or a fusion of different cost functions have been used in the literature as well [7]. other cost functions such as AD-Census [8] in which SAD and Census transform are used together or a combination of SAD and gradient function in [9] have been explored in the literature. In some of the techniques, the image is divided into textured and texture-less region for different cost functions and some of the techniques incorporate larger window size [10] for texture-less region and small window size near occlusions [11] . Several of these techniques use local adaptive window and different other cost functions which is not discussed further for brevity purpose. [12]

Local and global approaches have been studied well in literature. Global methods have also performed well in many cases. Belief propagation [13] uses Markov random fields to approximate minimum cost label in the energy function. In[14], the authors proposed an belief propagation, an effecient algorithm that uses a hierarchical method to reduce computation time and memory usage. In [15], the authors applied different alpha labels according to the locations of the local alpha expansion based on a MRF model with continuous label space. Dynamic programming has been used alongside different algorithms for scan line-based optimization fin the image. In it, the authors used RANSAC based method to detect occlusions and assign labels accordingly. Among the global approaches [16], patch match [17] based approach that uses iterative propagation and refinement from neighbouring pixels to obtain disparity value is also very popular [18].

Some other works use Minimum spanning tree (MST) [19] , Super-pixel based clustering (SLIC) [20] and iterative clustering algorithms to obtain disparity map and have better performance than the local techniques. Super pixel-based approach uses a free contour to retain the edge structure of objects, where pixels belonging to the same super pixel have a high likelihood to share the same disparity. However, with rise in use of deep learning approaches for different computer vision problems, they have also been used for disparity estimation and are discussed briefly in the following subsection.

## 3.2 Learning based depth estimation techniques

Most of the global stereo disparity methods discussed above have an inherent disadvantage of being computationally intensive and the non-global approaches are non-optimal. There is a significant amount of research for the use of deep learning architectures [21] to solve the bottlenecks in stereo disparity estimation problem. Training CNN for matching cost between image patches was initially introduced by Zbontar and LeCun [22] and at present there are lot of end to end networks for stereo matching and disparity estimation. These networks learn all steps of disparity computation together resulting in better outputs [23]. DispNet [24] is the first end-to end network along with a correlation network DispNetC for disparity. There are a lot of methods that only use 2D convolution and provide promising results like GwcNet [25], Stereo Transformer (STTR) [26], HITNet [27], AANet [28]. GwcNet presents an enhanced cost volume representation using a group wise correlation volume. There are several methods that use 3D convolutions for end-to-end stereo matching like GA-Net [29] that uses guided aggregation. GC-Net [30] and PSMNet [31] are methods that construct concatenation-based feature volume and use a 4D cost volume with 3D CNN to aggregate features. GC-net uses a 2D convolution neural network for dense feature representation while PSM-Net uses a stacked 3D convolution hourglass structure to aggregate cost volume. CRE Stereo is an end-to-end neural network that uses an adaptive correlation module to register locations in multi-scale feature space. It uses a stacked cascade architecture and works by down sampling the image pair before constructing the image pyramid. Recent methods like RAFT-Stereo [32] uses an iterative refinement module. It uses R-CNN for cost aggregation and works by down sampling to lower resolution, thus saving memory and computation resources.[33] CNN based techniques display a significant accuracy boost over the previous approaches. It is well known that 3D CNNs require more time for traiinig and the memory requirements is high in most cases, the cost construction mechanism of most of the network architectures is supported at low resolution thus putting a constraint by limiting their application. Moreover, most approaches are unable to handle occlusions explicitly, even in situations where the disparity in occluded regions can be an arbitrary value. Also, there is no explicit uniqueness constraint that can be imposed during the pixel or block matching, which brings a change to the performance due to inconsistencies. While most of these approaches work well with datasets like KITTI[34], Middlebury, KITTI, NYU Depth etc., The model trained by synthetic data has poor generalization and cannot be deployed in new scenarios. It is necessary to note that any model or architecture that's pre trained on a synthetic dataset cannot easily be generalised to an actual scene dataset due to heterogeneous nature data sources for real sequences [3]. Fine-tuning is necessary to enhance the transferability of learning-based methods, as ground truth data varies across different datasets.

By training on diverse datasets, learning-based methods can generalize well and infer disparities even in challenging texture less regions. They can learn to associate other visual cues, such as edges or gradients, with depth information. However, more challenges exist as the method need not only learn about image semantics but the actual depth too, which is a challenging task for aerial images.

# 4 Disparity estimation architecture

Disparity estimation architectures typically compute costs, aggregate them in multiple directions, and undergo refinement. These methods can be classified as local, global, or semi-global. Local algorithms choose the disparity with the lowest cost for each pixel, while global techniques minimize an energy function across the entire dataset. This section discusses the traditional semi-global block matching with its three cost variants, optimization-based Patch Match, and learning-based techniques, which are examined in this study.

## 4.1 Semi Global Block matching

Semi-Global Block Matching (SGBM) [35] is a stereo vision algorithm that offers low computational time and good accuracy in both static and dynamic situations. The key feature of the SGBM approach is the cost function to measure the disparity between corresponding pixels in stereo images. Costs are calculated for each pixel across all potential disparity levels which provide the initial disparity map. These cost values are stored in a row-major sequence, with neighboring pixel costs represented as a cost volume. Aggregation can be performed in a loop to process each path separately 【36】. 4, 8, or 16 paths, with the 4 or 8-path options typically used to keep computation time manageable. Cost aggregation function $E(d)$ can be defined as:

$$E(d) = \sum_{P} C(p, D_p) + \sum_{q \in N_p} P_1 T \vee D_p - D_q \vee \lnot = 1 + \sum_{q \in N_p} P_2 T |D_p - D_q| > 1 \lnot \quad (1)$$

In equation 1, $P_1$ is the constant penalty applied when disparity value changes by 1. $P_2$ is used when disparity value change is greater than 1.

The overall aggregated cost $(p,d)$ for $p$ is estimated by aggregating costs in $1D$ across all directions equally. The cost $L_r(p,d)$ for the path taken in $r$ direction is:

$$L_r(p,d) = C(p,d) + min(L_r(p-r,d), L_r(p-r,d-1) + P_1, L_r(p-r,d+1) + P_1, min L_r(p-r,i) + P_2) - min L_r(p-r,k) \quad (2)$$

The costs $L_r$ is then summed over all the 8 paths.

$$S(\mathbf{p},d) = \sum_{\mathbf{r}} L_\mathbf{r}(\mathbf{p},d)$$

In this equation, $L_r(p − r,d)$ is the aggregated value when the previous pixel disparity is d. $L_r(p − r, d − 1)$: represents the aggregation value when the disparity value of last pixel in the path is $d − 1$. $Min(L_r(p − r,I))$ represents the lowest value among all cost values from the path. $P_1$ and $P_2$ are input parameters which can be tuned as per the dataset. The total energy function in eq. (2) can be expressed in terms of a data and smoothness term. The disparity value belonging to the lowest aggregated cost is considered as the final disparity.

Errors arise in areas with invalid values, which can be corrected during post-processing steps such as peak removal and consistent disparity selection. These errors manifest as discrete disparity regions that significantly differ from their surroundings. Smaller patches are less likely to represent legitimate scene structures, allowing for the application of a size-based threshold to eliminate them. Various similarity measures, including gray value differences, census transform, and entropy, can be used to compute costs. The three functions employed in this study for aerial datasets are Birchfield-Tomasi (BT), Sum of Absolute Differences (SAD), and Adaptive Census (ADC).

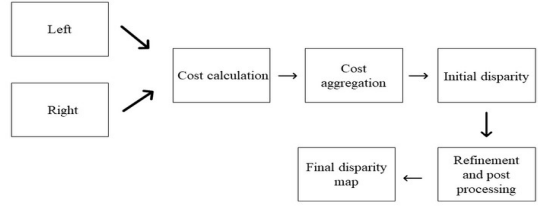

Figure 1: SGBM Pipeline

### 4.1.1 Birchfield-Tomasi dissimilarity

The Birchfield-Tomasi (BT) dissimilarity measure evaluates pixel similarity by assessing absolute variations in pixel intensities within a small area. A window is centered on the left pixel, and a horizontal scan line is drawn through the corresponding right pixel to calculate the BT dissimilarity between the current pixel and the epipolar line pixel in the right image. This measure is utilized for stereo matching in the OpenCV SGBM function and is defined as follows:

$$d_l(x_l, x_r) = \min_{x_r - \frac{1}{2} \leq x \leq x_r + \frac{1}{2}} \left| I_l(x_l) - \hat{I}_r(x) \right|$$
$$d_r(x_l, x_r) = \min_{x_l - \frac{1}{2} \leq x \leq x_l + \frac{1}{2}} \left| \hat{I}_l(x) - I_r(x_r) \right| \qquad (3)$$

Here $x_l$ and $x_r$ are the corresponding left and right pixel intensity values for left and right Images $I_l$ and $I_r$ respectively.

Instead of matching pixel to pixel, BT interpolates the pixel intensities and searches for half more pixel to get a more accurate match.

### 4.1.2 Sum of absolute difference

The Sum of Absolute Differences (SAD) is a pixel-based cost function that calculates the absolute differences in intensity values within a local window. The resulting sum reflects dissimilarity, with lower values indicating greater similarity. Although SAD is simple and computationally efficient, it is susceptible to variations in brightness and contrast between images, which can reduce its effectiveness in certain situations. SAD is defined as:

$$\sum(I_l(x,y), I_r(x+d,y)) = \sum |I_l(x,y) - I_r(x+d,y)| \qquad (4)$$

$x$ and $y$ are the coordinates and $d$ represents disparity.

### 4.1.3 AD-Census

This cost function is a combination of two cost functions the absolute difference and census transform, wherein the census transform is a binary descriptor that encodes the spatial relationships between pixels in a local window. The matching cost value between the two windows is computed using the Hamming distance between their census transforms. Hamming distance quantifies the number of positions at which the corresponding bits differ between two binary strings. For every pixel both the costs are calculated and added after normalizing them using their respective constants, resulting in one cumulative cost defined as:

$$C_{AD}(\mathbf{p}, d) = \frac{1}{3} \sum_{i=R,G,B} \left| I_i^{\text{Left}}(\mathbf{p}) - I_i^{Right}(\mathbf{pd}) \right| \quad (5)$$

$$C(\mathbf{p},d) = \rho(C_{\text{census}}(\mathbf{p},d), \lambda_{\text{census}}) + \rho(C_{AD}(\mathbf{p},d), \lambda_{AD}) \quad (6)$$

$$\rho(c, \lambda) = 1 - \exp\left(-\frac{c}{\lambda}\right) \quad (7)$$

here, $C_{AD}(p,d)$ is the cost function describing the absolute difference. $C_{AD}$ represents the cost value obtained through census transform and $C$ is the cumulative cost value wherein $\rho$ and $\lambda$ are the parameters determining weighting factor for both the costs.

## 4.2 Optimization based Patch Match

Among various optimization techniques for dense depth map estimation, the Patch Match scheme is chosen for analysis and comparison. The Patch Match algorithm identifies nearest matching patches between two images, employing random initialization [37], iterative propagation, and neighbor-based estimation. It computes correlations among neighboring pixels, iteratively sending costs to the next matching point. The random initialization generates a random disparity for a normalized plane, which is later merged into a plane equation, followed by iterative propagation aimed at reaching a global minimum. This paper presents results from a single iteration, noting that additional iterations can yield slightly improved outcomes. The results are post-processed using left-right consistency checks, weighted median filtering and occlusion filling. The process of Patch Match can be summarized as follows:

$$d_p = a_{f_p} p_x + b_{f_p} p_y + c_{f_p} \quad (8)$$

Here, $a_f$, $b_f$ and $c_f$ are the parameters of plane $f_p$ while $p_x$ and $p_y$ denote the pixel's $x$ and $y$ coordinates. The plane $f_p$ is of interest since the objective is to select the minimum aggregated cost among all the planes. That corresponds to the obtained disparity value.

$$f_P = argmin\, m(p,f) \quad (9)$$

In this method, the matching cost is calculated using the Hamming distance divided by the number of pixels in the window. While this computationally intensive approach produces visually appealing results and smooths fine edges, it has one of the slowest inference times among all methods, despite its high-quality output.

## 4.3 Deep Learning based approaches

Learning-based methods have achieved impressive results in stereo disparity estimation benchmarks but still require further improvement[21]. Supervised methods necessitate a substantial amount of annotated data with available ground truth disparity maps for training. These methods employ convolutional neural networks (CNNs) to learn the relationship between input stereo image pairs and their disparity maps. In this study, CRE Stereo, HitNet[27], and RAFT Stereo[32] have been used, utilizing their pre-trained models due to their high accuracy in Middlebury benchmarks.

### 4.3.1 CRE Stereo

In the Cascaded Recurrent Network with Adaptive Correlation, a pair of stereo images is input into feature extraction networks to create a 3-level feature pyramid. This pyramid facilitates correlation computation at various scales within a 3-stage cascaded recurrent network, starting at 1/16 of the input image resolution with all disparity values initialized to zero. For subsequent cascade levels, the upsampled prediction from the previous stage is used as the initialization for disparity refinement. The output from each stage is fed to the correlation layer for reducing the matching ambiguities in case of stereo image pairs not rectified or not on the epipolar line. A local feature attention mechanism matches points within a local window, complemented by an attention module before the correlation layer to aggregate global context. The 2D-1D alternate local search addresses non-rectified stereo pairs using a cost function that incorporates disparities from all pixels in both horizontal and vertical directions. Group-wise correlation is applied to the cost volume, ensuring the preservation of high-resolution details.

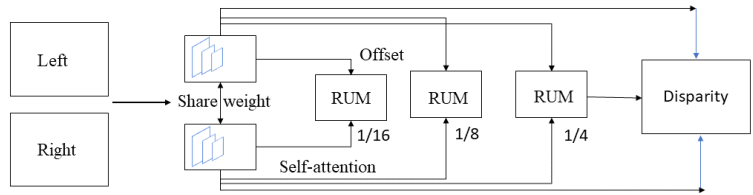

Figure 3: CRE Stereo pipeline

### 4.3.2 HITNet

HITNet is a neural network framework for depth estimation that balances real-time performance with accuracy. It decomposes the stereo matching problem into smaller overlapping tiles, processed hierarchically with iterative refinements. To address the computational challenges of 3D cost volumes, HITNet employs image warping, spatial propagation, and faster high-resolution initialization within its architecture.

HITNet adheres to traditional matching principles through three main steps: compact feature representation, high-resolution disparity initialization, and efficient propagation for refining estimates. Its feature extraction module employs a U-Net-like architecture. During initialization, matches for all disparities are computed exhaustively, but only the index location of the best match is stored during testing, eliminating the need to retain the entire cost volume. In the propagation step, the input consists of tile hypotheses, which are refined through spatial propagation and feature warping to predict accurate offsets. This step also predicts a confidence value to effectively fuse estimations from earlier layers and the initialization phase. This approach enables flexible learned representations and achieves strong results.

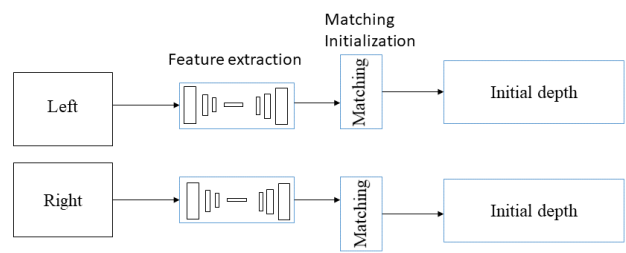

Figure 3: HitNet Pipeline

### 4.3.3 RAFT

RAFT constructs a 4D cost volume by extracting features from input images and calculating pixel correlations. The model consists of three main aspects: (1) an encoder that generates features for each pixel; (2) a 4D correlation volume for all pixel pairs; and (3) a recurrent GRU-based operator that refines disparity. The network employs blocks that downsample the input to produce feature maps at 1/4 and 1/8 of the original resolution. The cost volume is refined here using a series of 3D convolutions and mapped to a point-wise depth estimate. The context features and disparity along with correlation metrics are concatenated and fed into a hidden state, which predicts the disparity update. The final outputs are combined into a single feature map for further processing.Up sampling the obtained disparity to match ground truth disparities is recommended but it works by converting the input to low resolution. This is a very fast approach and yields convincing results.

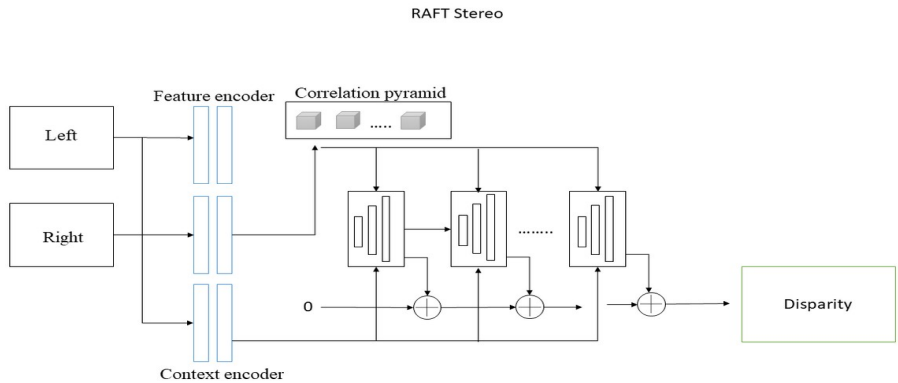

Figure 5: RAFT Stereo

## 5 Experimentation

In this section, the dense disparity estimation techniques for aerial images are compared quantitatively and qualitatively. The standard error metrics such as mean square error (MSE), SSIM and bad matched pixels (BMP) has been used to benchmark the performance of different algorithms used in this work. The overall pipeline for code development and testing was carried out on a linux system. The SGBM algorithm was implemented using the source code of OpenCV and the same was modified to compare the performance with different cost functions while keeping the remaining pipeline exactly same. The source codes for the CNN based architectures HITNet [27], CRE [38] , RAFT [39] has been taken from their respective repositories.

Along with this, an optimization-based approach of Patch match was also used for benchmarking its performance as compared to the non-optimization based techniques. It is hypothesized here that the performance of other optimization based approaches such as graph cuts, local alpha expansion would yield similar performance.

The images in WHU dataset is divided in three categories, that are, building, trees, and mixed region whereas for Mid Air dataset they are divided in trees, flat and rocky regions. A small proportion of images from each category is selected in random manner from these categories as depicted in Table 1.

**Table 1** Number of images taken from WHU and Mid Air datasets

| WHU Dataset | | | Mid Air Dataset | | | |
|---|---|---|---|---|---|---|
| Image type | Buildings | Trees | Mix | Flat | Rocky | Trees |
| Number of Images | 90 | 30 | 50 | 85 | 100 | 80 |

## 5.1 Visual analysis on WHU dataset

The WHU dataset, developed at Wuhan University, consists of 1,700 stereo images divided into training and testing subsets. Captured by the GF-7 satellite, these images represent diverse landscapes, including urban, rural, forested, and mountainous areas. The dataset includes ground truth disparity maps, making it ideal for evaluating stereo matching algorithms. Images from three categories have been tested using various disparity estimation schemes, as shown in Figure 1. 1(a) (b) and (c) represents the original left image as provided in the dataset, 1Row 2 (second row) corresponds to the ground truth disparity map provided by the publishers. The disparity images shown here are in gray scale and normalized on a common scale to maintain uniformity. In 1 (d), (e) and (f) the original left images from Mid-Air dataset are represented, similarly for subsequent Rows the disparity output of different techniques can be observed.

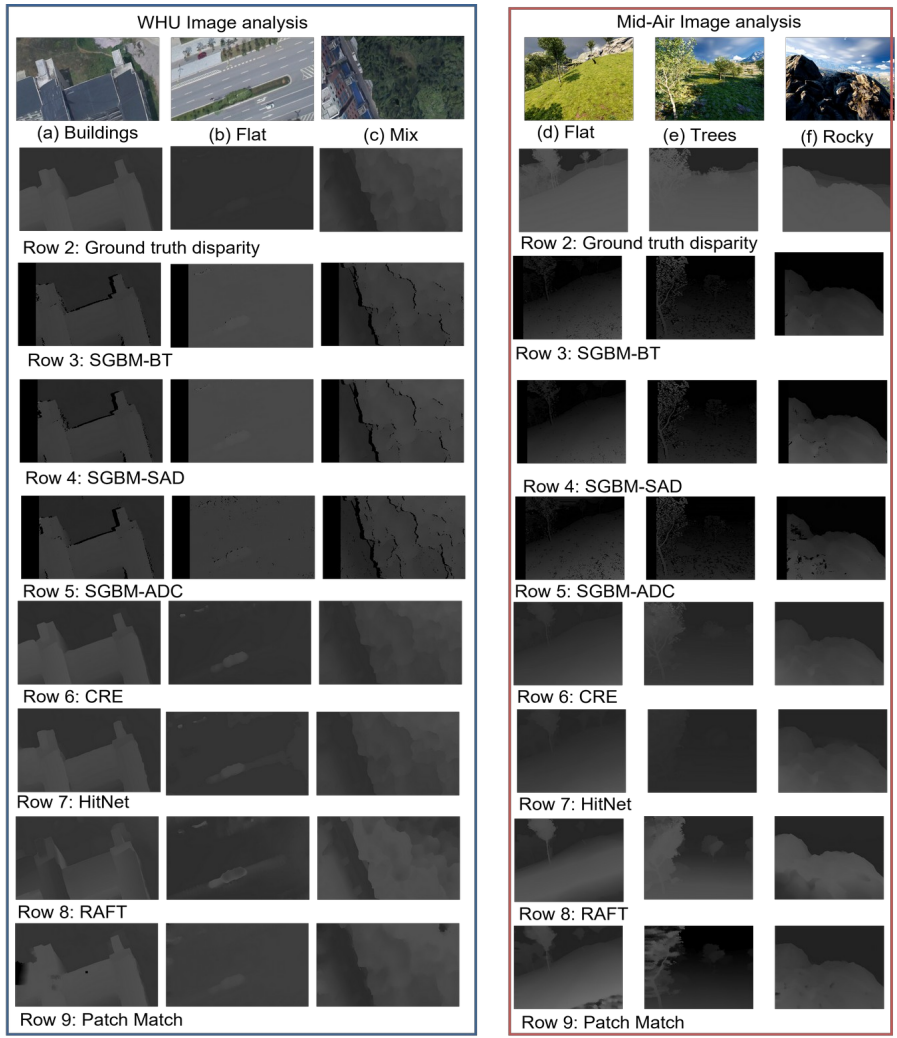

**Figure 6:** WHU and Mid Air dataset sample

The number of disparity levels has been configured to 96 for SGBM, which implies that for every pixel (*x*,*y*) in the left image, the algorithm calculates the cost for pixels ranging from (*x* − 96,*y*) to (*x*,*y*) in the right image. The number of disparities is a crucial parameter whose value if set too low can lead to a lot of noise and incorrect matches, especially in regions where the actual disparity is larger than the number of disparities set. Therefore, it is essential to choose an appropriate value for the number of disparities to achieve reliable stereo matching results.

It can be seen from Figure 1 that the SGBM does not yield as good result as much as the deep learning frameworks especially for images that are texture-less or contain too many discontinuities. The primary reason for this can be attributed to limited window size and local information available in window-based matching unlike learning based approach that uses global information. In case of image with well-defined structures, that is, buildings, represented as left most image column, the performance of SGBM is relatively better as compared to other images which have low texture areas. The zoomed in portion of the building region is shown separately in Figure 2 to compare the performance of different schemes. As seen in the zoomed in region, for ADC, SGBM-BT and SGBM-SAD, the edge information is not preserved to the same extent as that in SGBM – ADC as the latter is more robust over a window than the pixel based intensity difference. The black portions beside the edges can be attributed to the mis-matches. As compared to the SGBM based approach, the learning based methods show good performance, with RAFT preserving edge information of buildings better than HITNet for occluded cases. CRE provides information about the structure and perception of depth is better while the edges are smoothed out as compared to RAFT. Patch Match produces visually good results except for corners where it is not able to handle edges and occlusion very well as can be seen through a the Figure for the building image.

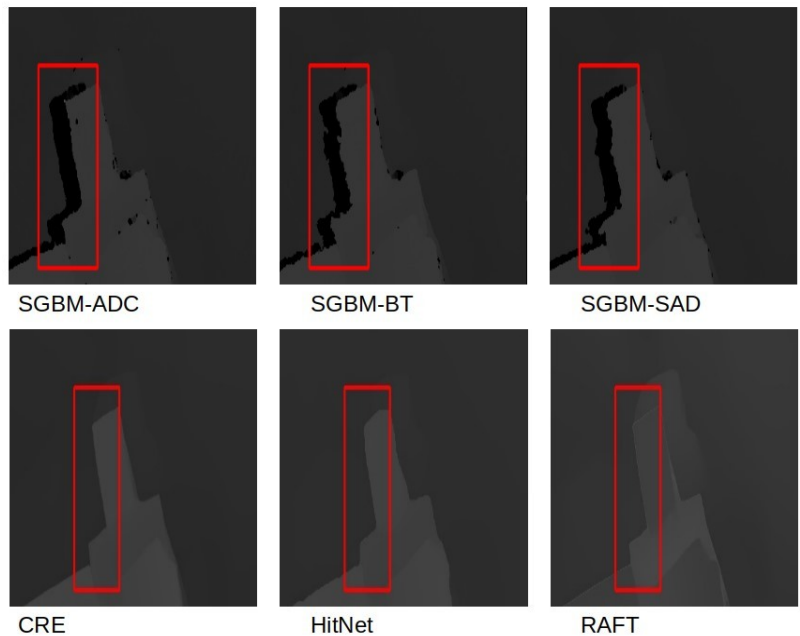

**Fig. 7** Analysis on zoomed building disparity output on WHU dataset

Figures 2 and 3 show that learning-based methods effectively detect edges and manage occlusions in flat regions, even in wide texture-less areas, as they are trained on diverse representations rather than relying solely on pixel intensity differences. In contrast, the SGBM algorithm struggles to find reliable matches due to the uniform color information of most pixels, resulting in blurring near occlusions and discontinuities. Improving SGBM performance could involve pre-processing, post-processing, parameter tuning, or using alternative algorithms. Patch Match outperforms other techniques due to its globally optimized results and consistency across various image types. In the third type of images, such as those containing buildings and trees, SGBM-ADC performs notably better due to the presence of varied textures, distinct features, and edges. In comparison, SGBM-BT and SGBM-SAD show inferior performance. SGBM's smoothness constraint helps manage depth variations at discontinuities, resulting in visually coherent disparity maps. However, learning-based techniques struggle with accuracy in certain areas, leading to poor edge definition for buildings. While CRE employs iterative refinement to capture fine details and depth in dense regions, it lacks the visual quality of other methods in utilizing contextual information and global constraints. RAFT, utilizing Recurrent Neural Networks, refines disparity using data from neighboring pixels, resulting in better overall performance compared to other learning approaches.

## 5.2 Visual analysis on Mid Air dataset

The Mid-Air dataset is an aerial stereo dataset that provides a large number of stereo images captured in a synthetic environment wherein the images are captured from a drone in different settings. It contains high-resolution stereo pairs with large baseline and wide field of view. It also provides accurate ground truth disparity, which is used for evaluating the accuracy of the algorithms. Including such datasets in evaluation not only helps in comparative analysis but also provides a benchmark to improve and enhance the quality of the algorithm. The overall dataset contains...images with different trajectories and weather conditions as seen visually. In order to test the efficacy of different disparity estimation schemes, three subsets of this dataset have been created by sampling images which has trees, rocky areas and mixed region as shown in Figure 1. This helps in understanding the performance of different techniques in different kinds of terrains and has been accordingly represented in Figure 3. The first row represents the original left image as captured by the UAV and the remaining images are arranged similarity as WHU dataset example.

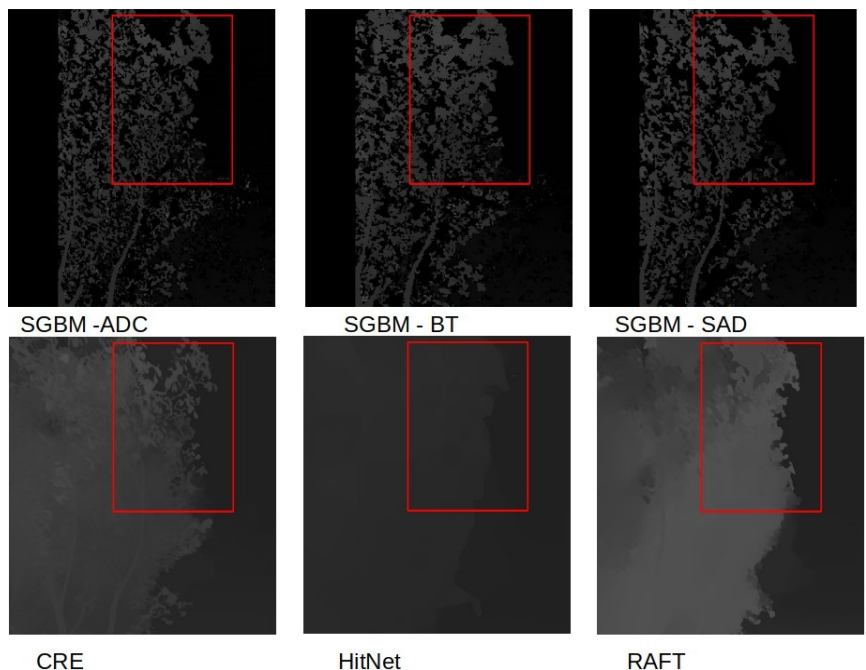

**Figure 8:** Analysis on zoomed trees disparity output on Mid-Air dataset

The disparity estimation results for images with trees favor SGBM variants over learning-based approaches. SGBM methods effectively distinguish tree leaves, unlike the learning models, which often lack structural detail compared to the original images. SGBM-BT produces a smooth output that merges background and foreground information, whereas SGBM-ADC excels in texture-less areas, offering superior ground detail compared to SGBM-BT and SGBM-SAD. Among learning-based methods, CRE-stereo outperforms others but still trails behind SGBM variants, as seen in the cropped leaf region. RAFT generates a disparity map that lacks sharpness, making leaves less distinct, while HITNet yields similar results, providing a rough indication of trees with smoothed edges. Patch Match, with its iterative plane refinement, offers clean and visually acceptable scene information, successfully differentiating flat regions from trees, but loses edge and texture detail in the process. In the Mid Air dataset, flat regions consist of large areas with similar textures and minimal depth variation. SGBM-ADC outperforms SGBM-BT and SGBM-SAD, clearly defining boundaries for nearby objects, though these details fade for distant objects. In contrast, learning-based methods struggle with both near and far objects; the ground often merges with the sky, making them indistinguishable. CRE and HITNet produce blurred outputs near edges, sometimes assigning similar disparities to the sky and flat ground, resulting in unwanted smoothing. RAFT, however, maintains clarity at boundaries. PatchMatch excels in texture-less regions, providing superior results compared to other methods.

Overall, visual analysis indicates that SGBM variants yield less accurate results in discontinuous depth and occluded regions compared to clearer areas. However, SGBM-ADC performs better near edges and retains some occluded regions. Learning-based approaches provide good estimates of image semantics but require further tuning to match the performance of optimization-based techniques. CRE and RAFT are effective in low-textured or repetitive surfaces, such as wood and grass, and handle occlusions well. HITNet occasionally lacks crucial details for specific applications, while CRE's iterative updates can lead to blurred images due to context loss during information propagation. PatchMatch consistently outperforms other methods but at a high computational cost. Therefore, developing faster algorithms that leverage computer vision geometry, rather than solely relying on learning-based architectures, is essential.

## 5.3 Quantitative analysis

This section employs various error metrics to compare the disparity maps generated by different SGBM variants, learning-based approaches, and the optimization-based PatchMatch technique. These metrics facilitate informed decisions regarding the performance of each method across different regions, such as flat, textured, and texture-less areas. The goal is to identify a technique that achieves promising results for various aerial images while balancing time, complexity, accuracy, and overall depth map quality.

Aerial images typically exhibit a larger disparity range compared to non-aerial images due to the diverse objects captured by top-down cameras, such as buildings, trees, and flat surfaces. Experimental results show that different techniques yield varying disparity ranges, necessitating both non-normalized and normalized comparisons for accurate evaluation. Moreover, comparing raw results offers deeper insights into each method's effectiveness, which is reflected in the metrics discussed in the subsequent subsections.

### 5.3.1 Mean Squared Error (MSE)

Mean squared error (MSE) is a simple and intuitive measure to quantify the difference between the ground truth disparity map and the obtained disparity map and is represented as:

$$MSE = \frac{1}{N} \sum_{i=1}^{N} (x_i - y_i)^2 \qquad (10)$$

In this equation, $x_i$ and $y_i$ are the images being compared with $N$ number of pixels. The errors are squared to amplify error values.

**Table 2** Comparison of MSE for WHU dataset using different disparity estimation techniques

|  | WHU Dataset Non-Normalised | | | | WHU Dataset Normalised | | | |
| --- | --- | --- | --- | --- | --- | --- | --- | --- |
| Disparity technique | Building | Flat | Mixed | Overall | Building | Flat | Mix | Overall |
| SGBM-ADC | **104.69** | 110.62 | **99.63** | 103.37 | 71.55 | 83.71 | **66.59** | 81.7 |
| SGBM-SAD | 109.166 | 106.51 | 104.79 | 103.95 | **66.54** | **80.85** | 67.16 | 70.5 |
| SGBM-BT | 105.97 | 117.83 | 105.98 | 106.21 | 72.78 | 83.36 | 65.87 | 71.34 |
| CRE | 113.57 | 107.05 | 116.8 | 108.86 | 102.25 | 99.01 | 103.2 | 95.56 |
| HITNet | 105.26 | **100.56** | 108.21 | 113.19 | 87.85 | 91.96 | 101.17 | 108.55 |
| RAFT | 109.35 | 106.23 | 102.62 | 104.14 | 104.94 | 99.85 | 100.88 | 101.95 |
| PM | 105.1 | 57.54 | 99.66 | 97.23 | 59.24 | 69.07 | 53.91 | 59.86 |

**Table 3** Comparison of MSE for Mid-Air dataset using different disparity estimation techniques

|  | Mid-Air Dataset Non-Normalised | | | | Mid-Air Dataset Normalised | | | |
| --- | --- | --- | --- | --- | --- | --- | --- | --- |
| Disparity technique | Trees | Flat | Rocky | Overall | Trees | Flat | Rocky | Overall |
| SGBM-ADC | 126.2 | 130.26 | 138.89 | 121.59 | 108.12 | 108.84 | 95.45 | 107.86 |
| SGBM-SAD | 122.06 | 130.16 | 136.43 | 122.76 | 107.27 | 118.64 | 85.65 | 105.91 |
| SGBM-BT | 121.03 | 131.75 | 139.73 | 123.64 | 106.93 | 93.45 | 89.15 | 106.89 |
| CRE | 90.86 | **79.62** | **81.12** | **89.17** | 85.05 | 86.98 | **74.12** | **79.71** |
| HITNet | **86.33** | 88.62 | 84.27 | 90.77 | 88.81 | **85.12** | 80.46 | 86.68 |
| RAFT | 110.72 | 115.94 | 119.71 | 112.3 | 104.15 | 110.02 | 85.73 | 107.23 |
| PM | 89.34 | 87.26 | 84.19 | 87.27 | 86.68 | 84.46 | 79.79 | 84.09 |

Table 2 and Table 3 compare the MSE value for disparity obtained from different techniques in two different datasets, that is, WHU and Mid-Air dataset considered here. In case of WHU and Mid-Air dataset, the MSE value between the ground truth and non-normalized disparity value is found to be very high as compared to the normalized cases. A normalization of the disparity value is carried out in the range of 0 -75 , which is chosen as the median of minimum and maximum disparity value in the ground truth images for WHU dataset. In case of Mid-Air data set the upper range value was 81 but has been chosen as 75 here for sake of simplicity.

The inferences from this table and are listed as follows:

1. MSE values for different variants of SGBM are similar for WHU and Mid-Air dataset and are lower among all for the normalized cases as compared to nonnormalized case.
2. MSE values for all the three learning-based techniques are relatively better than SGBM variants in texture-less and synthetic images.
3. PatchMatch as a technique performs better across all the different kinds of images and has better MSE than all the other methods. This can be attributed to the solution of disparity value obtained by solving an optimization function.
4. In case of mixed regions and regions with building, SGBM-ADC provides better results than SGBM-BT and SGBM-SAD, and other learning-based approaches in case of non-normalization.

Overall, it can be summarised that for flat regions, results of SGBM are not very good, since these regions consist of similar texture, colour information and no distinct edges or occlusions.

SGBM variants perform equally good for images with good texture such as building and mixed regions. The SGBM based approach work good for real world images in WHU dataset while the learning-based approach works better for synthetic images contained in Mid-Air dataset. However, MSE has certain limitations that is, it is sensitive to small variations and does not consider perceptual quality of the constructed disparity map.

### 5.3.2 Bad Matching Pixels (BMP)

It refers to the total count of pixels in an image that do not have a corresponding pixel in the other image or have a corresponding pixel that is significantly different in terms of colour or texture. Bad matching pixels can occur for several reasons, including occlusions, specular reflections, texture-less regions, and errors in the stereo matching algorithm. This metric has been very commonly used to measure the performance of disparity estimation algorithms and is explored here to compare different traditional and learning approaches. Bad matched pixel is defined as:

$$B(i,j) = \begin{cases} 1, \wedge if \, |D(i,j) - G(I,j)| > t \\ 0, \wedge otherwise \end{cases} \quad (11)$$

Here $D_{(i,j)}$ is the obtained disparity map and $G_{(i,j)}$ refers to the ground truth. $B_{(i,j)}$ is the bad matched pixel at location $(i,j)$. The percentage of bad matching pixels in the disparity map is calculated as:

$$P = \frac{1}{N} \sum_{i=1}^{h} \sum_{j=1}^{w} B(i.j) \times 100 \quad (12)$$

As seen in Figure 4, the BMP values for both the WHU dataset and Mid-Air dataset is found to decrease as the threshold value increases. In case of WHU dataset which has real images from UAV, the traditional SGBM based approach is found to have relatively higher error percentage as compared to the learning-based approaches. The performance of RAFT is comparable to other approaches and the performance of SGBM − ADC is better than other SGBM variants. The performance of Patch match is better than all other techniques owing to its inherent nature of finding a global minimum using optimization approach. In case of Mid-Air dataset with synthetic images, the performance of SGBM variants is similar and the reason for very high error percentage is that the obtained disparity values are widely different from the ground truth of synthetic images.

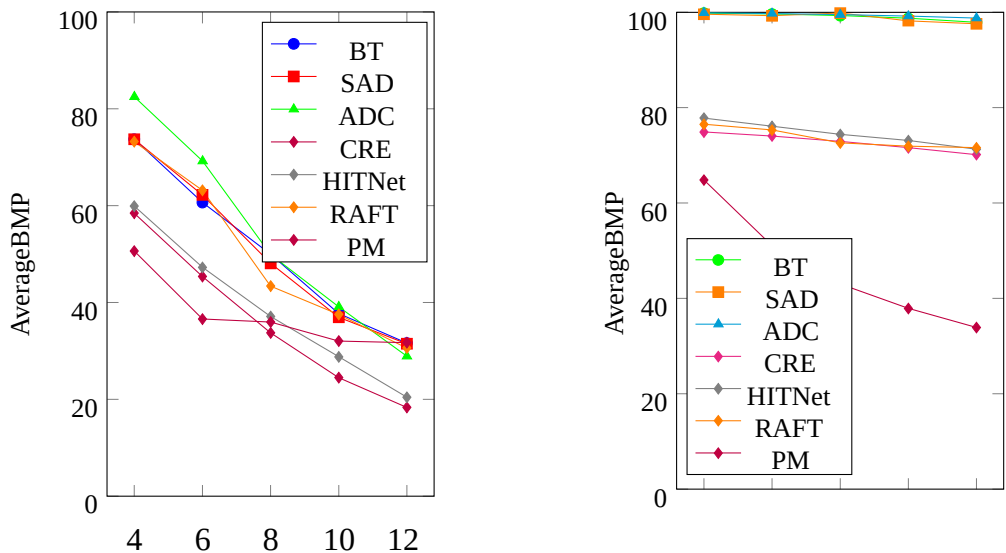

BMP Values for WHU dataset    4    6    8    10    12
BMP Values for Mid-Air dataset

### 5.3.3 Computation Time

The computation time of any algorithm is one of the major factors governing usage in real time application to compare the different methods experimented here. The time required to run different techniques on the two datasets for a single image pair is observed and plotted in Figure 5.3.4. In case of SGBM variants, it was found that changing cost functions led to minor change in computational time. The optimization-based approach PatchMatch used here takes relatively much larger time than the traditional SGBM based or learning based approaches. It is for this very reason that despite yielding good results, they are not practically realizable for real time systems. The time taken for Mid-Air dataset is higher as the image resolution is higher than the WHU dataset. These implementations are not optimized for parallel processing or to be run on optimized hardware resources which could have led to even lesser computation time in each of the cases.

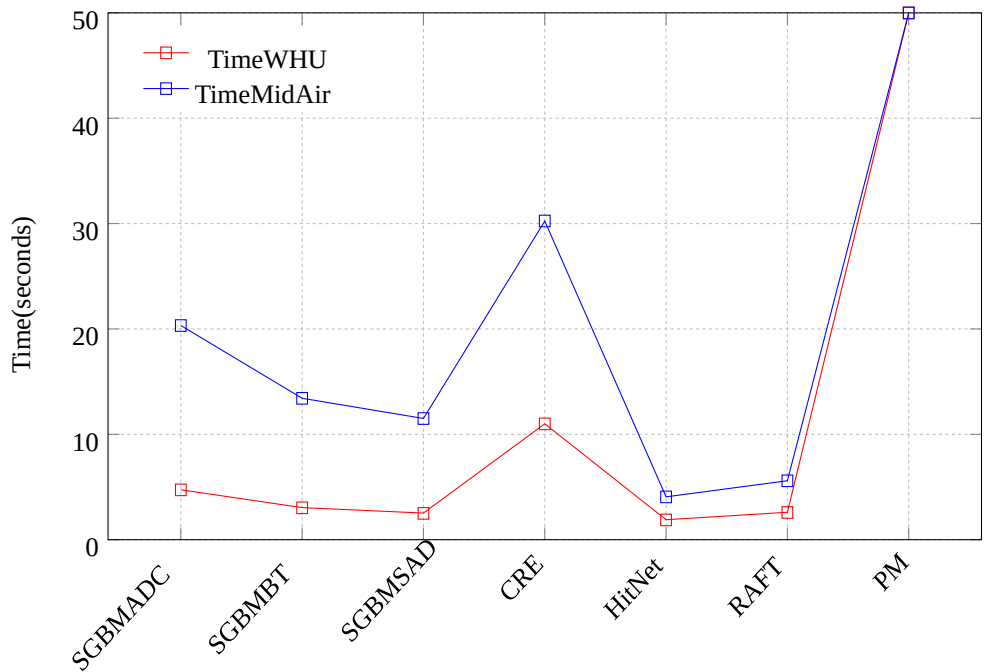

Figure 9: Computation time for different methods

### 5.3.4 Structural Similarity Index (SSIM)

The Structural Similarity Index (SSIM) is a metric for measuring similarity between two images, it's defined as:

$$SSIM(x,y) = [l(x,y)]^{\alpha}[c(x,y)]^{\beta}[s(x,y)]^{\gamma} \qquad (13)$$

where $x$ and $y$ are the two images being compared and are the constants that control relative importance of these three terms. The terms $l(x,y)$, $c(x,y)$, and $s(x,y)$ are the local luminance, contrast, and structural similarities, respectively, which are computed over a small window of pixels. SSIM values range between 0 and 1, where a value of 1 indicates perfect similarity and a value of 0 indicates dissimilarity. Figure 5 compares the performance of different disparity estimation techniques in both the WHU and Mid-Air dataset. The SSIM value for all the schemes is higher in case of WHU dataset as compared to Mid − Air dataset. The traditional SGBM based approach perform convincingly as compared to the learning-based approach in case of WHU dataset while the same is not the case in Mid-Air dataset. In case of Mid-Air dataset, the synthetic images have wide texture-less areas such as trees, ground, etc. which are known to create issues in block matching kind of techniques. However, for learning based techniques, these images are processed in different scales and have an output which is smoothed out such that there are no sudden changes in disparity levels. The SGBM based techniques can preserve the minute details to match the intensity levels in left and right image while reducing the overall structural information.

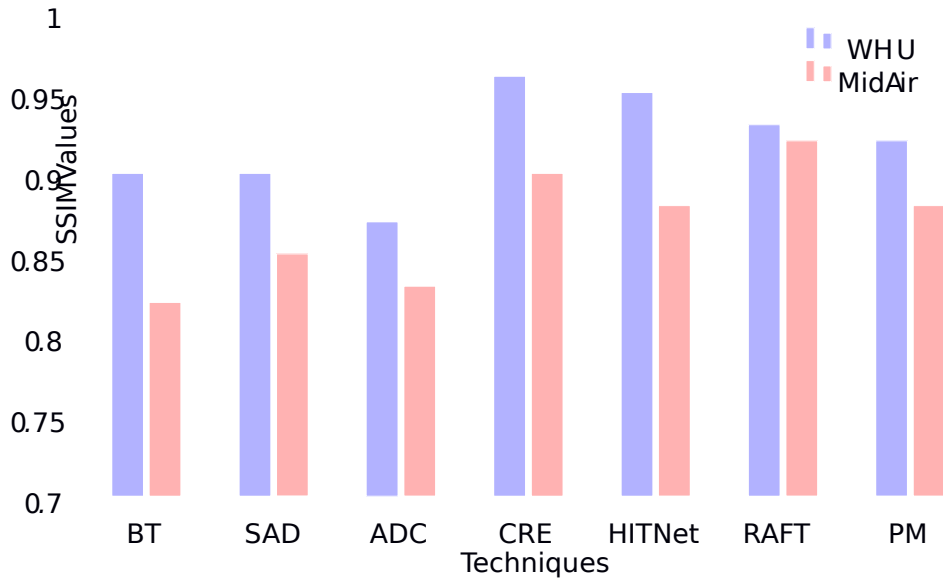

**Fig. 5** Comparison of SSIM values for different stereo disparity estimation methods

# 6 Summary

In this work various metrics to compare disparity maps obtained from different traditional and learning based approaches have been used. It was observed that image texture and features significantly influence the estimated disparity. The block matching method relies on low-level image features to find corresponding pixels in the left and right images, resulting in noisy disparity maps and reduced performance. However, despite this limitation, the method still produces disparity maps that preserve geometric accuracy here and their performance was nearly same different cost functions were explored.

In learning-based approaches, semantic segmentation is performed effectively. However, their MSE values differ widely from the ground truth image as they do not have any common reference. The dense depth map in case of SGBM based techniques are obtained by calculating the shift in disparity between left and right image, while in case of learning based approach, these values are based on a pre-trained model. Therefore, when the same approach is applied on a new kind of dataset, normalization of the result within a specified disparity level is very necessary. The learning-based approach gives a very good approximation of the relative disparity and if a hybrid approach is created to fuse the computational geometry like SGBM, it can yield far better results than either of the two approaches.

Among the learning based methods, CRE Stereo was found to be good and robust algorithm that performed well in almost all cases. It is memory efficient as it uses local search window instead of full cost volume. HITNet avoids full cost volume computation and adapts a coarse to fine propagation. Raft Stereo is fast, accurate and provides good results but blurs textures and occlusions. In the context of aerial images, the choice of technique for disparity depends on various factors like quality of images, scene complexity and colour or gradient information. It should be noted that different methods can be tuned to specific application or a particular dataset and for different regions or speed. However more efforts are required towards developing algorithms that are both geometrically consistent and fast.